\documentclass[conference]{IEEEtran}
\IEEEoverridecommandlockouts

\usepackage{cite}
\usepackage{amsmath,amssymb,amsfonts}
\usepackage{algorithmic}
\usepackage{graphicx}
\usepackage{textcomp}
\usepackage{subcaption}
\usepackage{xcolor}
\usepackage{multirow}
\usepackage{array}
\usepackage{url}
\usepackage[T1]{fontenc}   
\usepackage[utf8]{inputenc}
\usepackage{textcomp}      
\def\BibTeX{{\rm B\kern-.05em{\sc i\kern-.025em b}\kern-.08em
    T\kern-.1667em\lower.7ex\hbox{E}\kern-.125emX}}
\begin{document}

\title{Automated Mosaic Tesserae Segmentation via Deep Learning Techniques\\
}

\author{
\IEEEauthorblockN{Charilaos Kapelonis\IEEEauthorrefmark{1}, Marios Antonakakis\IEEEauthorrefmark{1}, Konstantinos Politof\IEEEauthorrefmark{1}, 
Aristomenis Antoniadis\IEEEauthorrefmark{2}, Michalis Zervakis\IEEEauthorrefmark{1}}
\IEEEauthorblockA{\IEEEauthorrefmark{1}School of Electrical and Computer Engineering, Technical University of Crete, Chania, Crete, Greece}
\IEEEauthorblockA{\IEEEauthorrefmark{2}School of Production Engineering and Management, Technical University of Crete, Chania, Crete, Greece}
\IEEEauthorblockA{Emails: \{ckapelonis, mantonakakis, kpolitof, aantoniadis, mzervakis\}@tuc.gr}
}

\maketitle

\begin{abstract}
Art is widely recognized as a reflection of civilization and mosaics represent an important part of cultural heritage. Mosaics are an ancient art form created by arranging small pieces, called tesserae, on a surface using adhesive. Due to their age and fragility, they are prone to damage, highlighting the need for digital preservation. This paper addresses the problem of digitizing mosaics by segmenting the tesserae to separate them from the background within the broader field of Image Segmentation in Computer Vision. We propose a method leveraging Segment Anything Model 2 (SAM~2) by Meta AI, a foundation model that outperforms most conventional segmentation models, to automatically segment mosaics. Due to the limited open datasets in the field, we also create an annotated dataset of mosaic images to fine-tune and evaluate the model. Quantitative evaluation on our testing dataset shows notable improvements compared to the baseline SAM~2 model, with Intersection over Union increasing from 89.00\% to 91.02\% and Recall from 92.12\% to 95.89\%. Additionally, on a benchmark proposed by a prior approach, our model achieves an F-measure 3\% higher than previous methods and reduces the error in the absolute difference between predicted and actual tesserae from 0.20 to just 0.02. The notable performance of the fine-tuned SAM~2 model together with the newly annotated dataset can pave the way for real-time segmentation of mosaic images. \\
\end{abstract}

\section{Introduction}

Cultural heritage is a reflection of society and preserving it is important for cultural diversity, growing a sense of belonging and imparting knowledge to the future generations \cite{stephenson2023}. Artworks are tangible expressions of cultural heritage and mosaics are a specific form of this art. Mosaics consist of small tiles (called tesserae) which are glued together to form 2D or even 3D patterns \cite{helenmiles2025}. This art form has existed for thousands of years and many modern artists continue to adapt this style. The long history of this art style highlights the importance of documenting and preserving these artworks \cite{getty2025}. Due to their particular structure and materials, mosaics are highly vulnerable to deterioration over time, often suffering catastrophic damage \cite{britannica2025}. For the reasons above, the digitization of mosaics is a very crucial task. This paper addresses the problem of automatically segmenting mosaic images into their individual tesserae.

Mosaic segmentation is a challenging task, as mosaics are often damaged over time and the tesserae could appear broken or partially unrecognizable. This problem falls within the domain of Image Segmentation, a key area of Computer Vision, which involves dividing an image into meaningful and non-overlapping regions of interest \cite{minaee2020}. The goal is to develop a method that takes a mosaic image as input and produces a binary segmentation mask in which the tesserae are clearly separated from the background.

Research on mosaic image segmentation remains limited, with traditional expert manual annotation being accurate but too slow for large-scale use. In 2004, an analysis of medieval mosaic conservation to reveal original patterns was presented in \cite{Zitova2004}, while the works \cite{Zarghili2001}, \cite{Zarghili2008} proposed a retrieval system for Arabo-Moresque decor images based on mosaic representations. In 2008, a method based on the Watershed Algorithm and k-means classification (yielding higher accuracy than a pixel-based strategy) was proposed \cite{WT2008}. In 2016, a group of researchers proposed Deformable Models using Genetic Algorithms, in which each possible segmentation is represented as a set of quadrangles whose shapes and positions change, aiming to capture variations in mosaics' age and style \cite{deformable2017}. In 2020, a Deep Learning (DL) technique based on a Convolutional Neural Network (U-Net) was proposed \cite{unet2020}, while in 2021, the Mo.Se. (Mosaic Segmentation) Algorithm was proposed, combining learning-based methods with procedural ones \cite{mose2021}. However, these methods still struggle with varied mosaic styles and often fail to handle damaged or incomplete tesserae, while relying on heavy computational resources.

This work focuses on building a tool capable of automatically segmenting mosaic images. The foundation of this project lies upon utilizing the advancements in Machine Learning (ML) and the field of Image Segmentation. We propose leveraging the latest foundation model of Meta AI, known as Segment Anything Model 2 (SAM~2) \cite{sam2}, which showcases impressive results in the task of segmentation. We evaluate the performance of our model and compare it to previous related works. This paper also involves a manually annotated set of mosaic images due to the lack of openly available images, thereby creating a state-of-the-art dataset, and fine-tuning SAM~2 by Meta AI. We, finally, analyze the specific configurations for training and evaluating the models, while also providing the results and comparisons to previous studies.

\vfill

\section{Material and Data}

\subsection{Overview}
Our proposed method involves fine-tuning SAM~2 by Meta AI \cite{sam2} in order to automatically segment mosaic images and extract the individual tesserae. SAM~2 is a foundation model with strong zero-shot segmentation capabilities \cite{zhu2024medicalsam2}, making it well-suited for adapting to domain-specific tasks with limited labeled data \cite{jiaxing2025}. 

\subsection{Data}

\subsubsection{Acquisition}
\begin{figure}[tb]
    \centerline{
        \begin{subfigure}[b]{0.25\textwidth}
            \includegraphics[width=\textwidth]{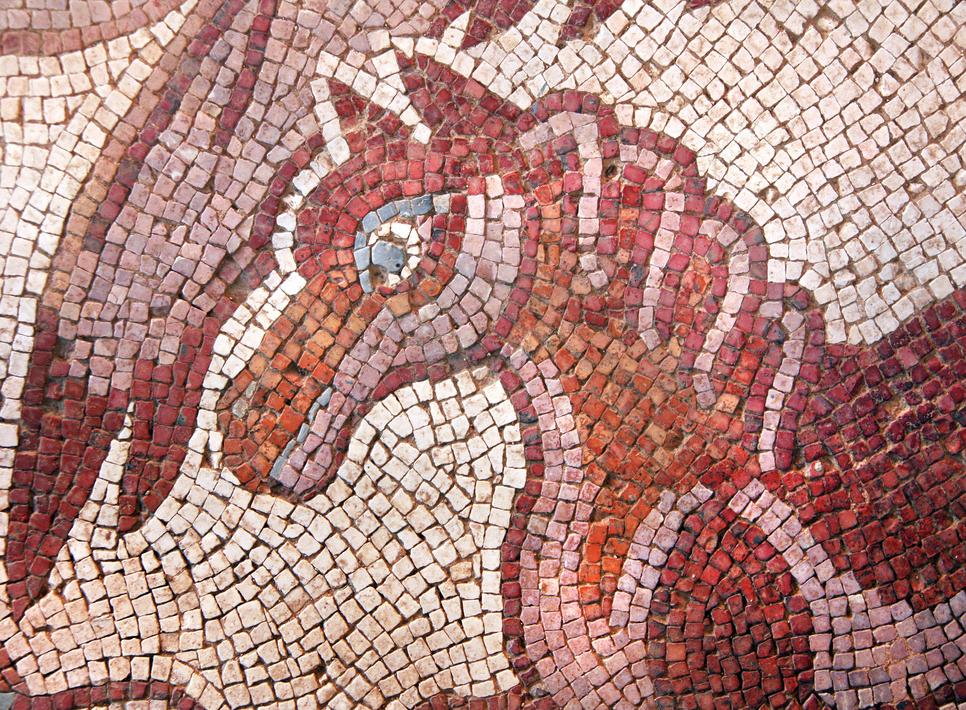}
            \caption{Original image}
            \label{fig:original}
        \end{subfigure}
        \hfill
        \begin{subfigure}[b]{0.25\textwidth}
            \includegraphics[width=\textwidth]{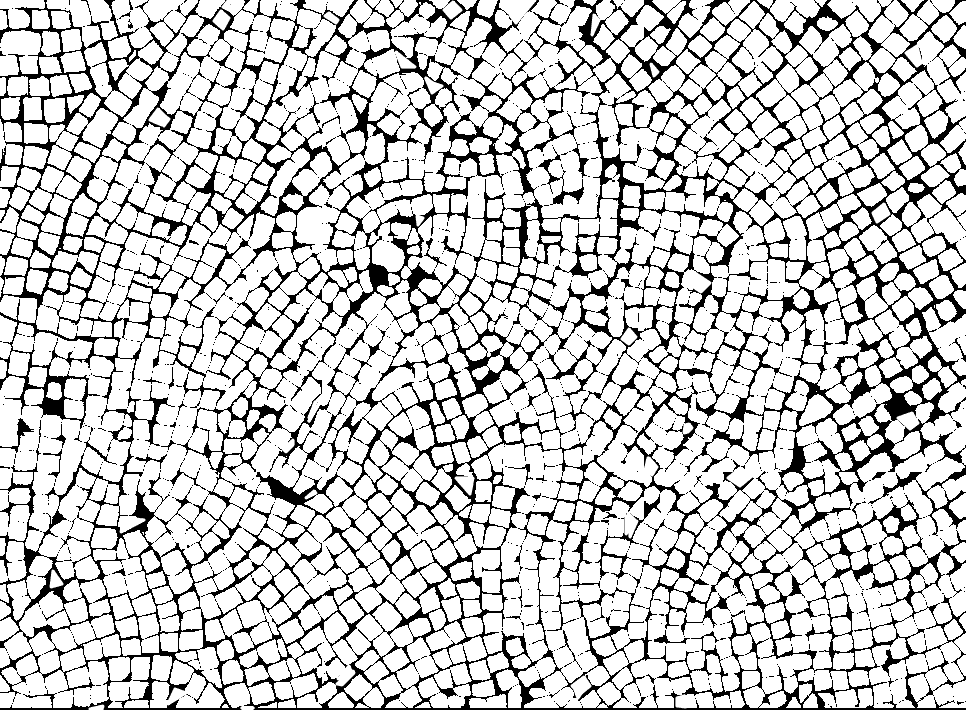}
            \caption{Binary mask}
            \label{fig:mask}
        \end{subfigure}
    }
    \caption{Example of an annotated image: (a) original image, (b) binary mask.}
    \label{fig:mosaic_and_mask}
\end{figure}

In cultural heritage tasks such as mosaic analysis, annotated data is often scarce or entirely unavailable. We retrieved our images from various public sources\footnote{Images sourced from stock image providers, e.g., iStock by Getty Images (\url{https://www.istockphoto.com}) and Adobe Stock (\url{https://stock.adobe.com}), used under their respective licenses.} thus creating our own dataset of 19 mosaics (e.g., see Fig.~\ref{fig:original}) and making it public\footnote{\url{https://huggingface.co/datasets/ckapelonis02/mosaics-images}}.

\subsubsection{Annotation}
The data annotation process includes retrieving mosaic images from public sources and producing the regions of interest, namely the tesserae, which will be referred to as masks. These masks were aggregated into a different image, which served as the ground truth image. To perform the annotation we operated with Label Studio\footnote{\url{https://labelstud.io/}}, which is an open-source Data Labeling platform. In this way, the annotation procedure became efficient and the masks more accurate. Given that SAM~2 is a promptable model, a single input point was required to guide the generation of each mask and was automatically created by computing the centroid of each annotated region.

\subsubsection{Augmentation}
\begin{figure}[tb]
    \centering
    \begin{subfigure}[b]{0.3\linewidth}
        \includegraphics[width=\linewidth]{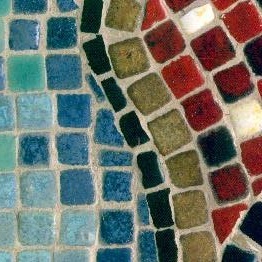}
        \caption{Original}
    \end{subfigure}
    \hfill
    \begin{subfigure}[b]{0.3\linewidth}
        \includegraphics[width=\linewidth]{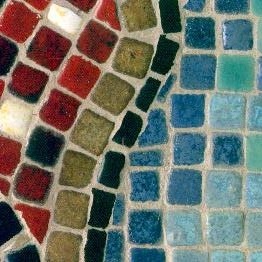}
        \caption{Horizontal Flip}
    \end{subfigure}
    \hfill
    \begin{subfigure}[b]{0.3\linewidth}
        \includegraphics[width=\linewidth]{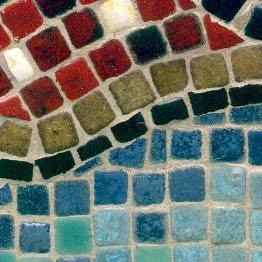}
        \caption{Rotation 270°}
    \end{subfigure}
    \caption{Augmentations applied to a single example image (cropped).}
    \label{fig:augmentations_470}
\end{figure}

The annotation process required a considerable amount of time because of the nature of the problem; most tesserae have complex shapes and appear damaged. To tackle this difficulty we employed several image augmentation techniques to further expand the annotated dataset. This process also aimed to increase data diversity, reduce overfitting, and allowed the model to become more robust to variations in tesserae appearance and mosaic structures. As shown in Fig.~\ref{fig:augmentations_470}, the images were first systematically cropped and then augmented through rotations of $90^\circ$, $180^\circ$, and $270^\circ$, as well as vertical and horizontal flips. These transformations preserve the semantic content of the data while introducing variations in the orientation, which is vital as the tesserae can appear at arbitrary angles.

\subsubsection{Data Splits}
The train dataset consists of 984 mosaics, for which we employed an 80\%-20\% split between the pure training images and the validation dataset, yielding 787 training and 197 validation images. The test dataset consisted of 166 cropped mosaic images. The resulting proportions are approximately 68\% for training, 17\% for validation, and 15\% for testing.

\subsection{Fine-Tuning}
We fine-tune the baseline \texttt{sam2\_hiera\_large} model\footnote{\url{https://huggingface.co/facebook/sam2-hiera-large}}, which offers the best segmentation performance among the available options, despite its larger size and slower inference speed. The SAM~2 architecture consists of three basic components: the image encoder, the prompt encoder and the mask decoder. The image encoder processes the images and produces the image embeddings, thus is the largest component and training it is infeasible with the available computational resources. We choose to train only the prompt encoder and mask decoder, while freezing the image encoder layer. We, finally, compare the performance of the fine-tuned model (\texttt{SAM2\_ft}) with the baseline model (\texttt{SAM2\_base}).

\subsection{Training}
\begin{figure}[tb]
\centerline{
    \includegraphics[width=0.20\textwidth]{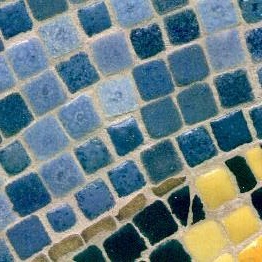}
    \hfill
    \includegraphics[width=0.20\textwidth]{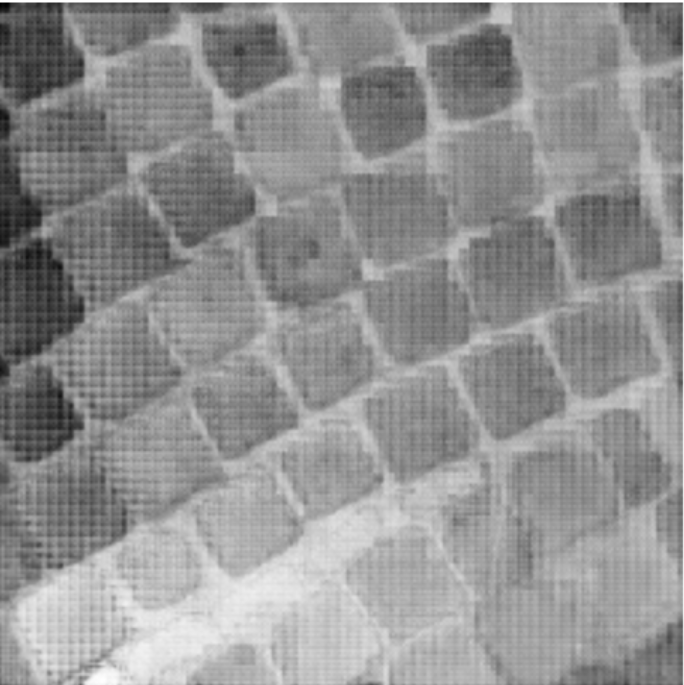}
}
\caption{(Left) A cropped, augmented training image. (Right) The corresponding raw mask logits, showing model confidence per pixel regarding tesserae classification.}
\label{fig:logits}
\end{figure}

For the training process, we utilize the SAM~2 Image Predictor, which is the component that predicts the tesserae based on the original image and the respective input points. More specifically, the model processes a batch of images, the size of which is a tunable parameter. For an image of the batch the prompt encoder creates sparse and dense embeddings, which in turn are used by the mask decoder to construct the predicted masks in low resolution. Every pixel is assigned to a raw prediction score, also referred to as logit, which indicates how confident the model is that it belongs to a tessera (Fig.~\ref{fig:logits}). The sigmoid function is eventually applied to the logits to transform the real numbers into probabilities.

\subsubsection{Loss Function}
Loss functions play a vital role in determining the model's performance. For sophisticated objectives such as segmentation, it is not possible to decide on a universal loss function \cite{jadon2020}. To enhance the model's performance for our specific task of segmenting mosaics, we are designing a loss function based on the Dice Coefficient, which is differentiable, to directly optimize the overlap between the predicted and the ground truth image \cite{ma2020}, \cite{sudre2017}.

Let $P$ be the predicted mask and $G$ the ground truth mask, both with pixel values normalized between 0 and 1.

The Dice Loss is defined as:

\begin{equation}
\mathcal{L}_{\text{Dice}} = 1 - \frac{2 \cdot (P \cdot G)}{\|P\|_1 + \|G\|_1}
\end{equation}

To further improve the model's performance, we also consider the model's confidence in its prediction by comparing the predicted confidence score to the actual overlap between the predicted and ground truth masks to find misclassification errors by rejecting low-confidence predictions \cite{zhu2023}. This method is widely known as calibrated confidence.

The predicted mask is thresholded at 0.5 to create a binary mask $\hat{P} = \mathbb{1}_{P > 0.5}$, where pixels with predicted values greater than 0.5 are considered foreground.

The Intersection over Union (IoU) between the ground truth $G$ and the thresholded prediction $\hat{P}$ is:

\begin{equation}
\text{IoU} = \frac{|G \cap \hat{P}|}{|G \cup \hat{P}|}
\end{equation}

The model also outputs a predicted confidence score $s$ for the mask, which ideally should reflect this IoU value.

The Score Loss penalizes the absolute difference between this predicted confidence score and the actual IoU:

\begin{equation}
\mathcal{L}_{\text{Score}} = |s - \text{IoU}|
\end{equation}

Finally, the total loss function combines the Dice Loss and Score Loss with a weighting factor $\lambda$:

\begin{equation}
\mathcal{L}_{\text{Total}} = \mathcal{L}_{\text{Dice}} + \lambda \cdot \mathcal{L}_{\text{Score}}
\end{equation}

\subsubsection{Validation}
We validate the updated model on a separate set to track performance and select the best checkpoint. Unlike training, we use SAM~2’s Automatic Mask Generator, which produces masks by generating a grid of points internally, thereby reflecting real inference conditions.

\subsubsection{Hyperparameter Tuning}
The training process involves tuning several critical hyperparameters to ensure effective model convergence and optimal performance. In this implementation, we employ the AdamW optimizer. The AdamW optimizer relies primarily on two key hyperparameters: the learning rate and the weight decay coefficient. In addition to these optimizer-specific parameters, we also tune the batch size, which denotes the number of training samples processed simultaneously in one iteration \cite{kandel2020}.

\subsubsection{Learning Rate}
The learning process heavily depends on choosing a suitable learning rate. To get the optimal results, a common practice involves using a learning rate decay, in which the rate is a monotonically decreasing function of the epoch iteration.

The learning rate $\eta_t$ at epoch $t$ is defined by the cosine annealing schedule \cite{loshchilov2017}:

\begin{equation}
\eta_t = \eta_{\min} + \frac{1}{2} (\eta_{\max} - \eta_{\min}) \left(1 + \cos\left(\frac{t \pi}{T_{\max}}\right)\right)
\end{equation}

where:
\begin{itemize}
    \item $\eta_{\max}$ is the initial learning rate
    \item $\eta_{\min}$ is the minimum learning rate
    \item $T_{\max}$ is the total number of epochs
    \item $t$ is the current epoch
\end{itemize}

The goal of this scheduling strategy is to allow large updates in the early stages for faster convergence, and smaller updates later to fine-tune the model weights.

\subsubsection{Weight Decay}
Weight decay is a regularization technique to prevent overfitting, enhancing the generalization ability of the trained model, by penalizing large weights during training. In our implementation, we adopt the decoupled weight decay strategy introduced by Loshchilov and Hutter \cite{loshchilov2019}, where weight decay is applied separately from the gradient-based update step.

In mathematical terms, the total loss function is calculated as:

\begin{equation}
\mathcal{L}_{\text{Total}} = \mathcal{L}_{\text{Original}} + \lambda \lVert \theta \rVert_2^2
\end{equation}

where:
\begin{itemize}
    \item $\mathcal{L}_{\text{Original}}$ is the original loss function
    \item $\theta$ represents the model parameters
    \item $\lambda$ is the weight decay coefficient
\end{itemize}

\subsubsection{Batch Size}
The batch size, which is the number of training records used in one forward and backward pass of the network, is a vital parameter of the training process. It is important to choose an optimal value for the batch size to balance the computational efficiency and memory constraints, while also maximizing the model generalization.

\subsubsection{Configuration}
We systematically experimented with various combinations of key adjustable hyperparameters, namely the learning rate, weight decay, and batch size, in order to identify an optimal configuration that balances effective learning with model regularization, while also considering computational efficiency.

More specifically, we experimented with the different combinations of the values below, conducting 12 runs in total:
\begin{itemize}
    \item Learning Rate: \{$10^{-5}$, 2$\cdot10^{-5}$, 3$\cdot10^{-5}$\}
    \item Weight Decay: \{$10^{-4}$, $10^{-5}$\}
    \item Batch Size: \{4, 8\}
\end{itemize}

\subsubsection{Metrics}
In image segmentation tasks, the performance of a model is commonly evaluated using metrics that quantify the overlap between the predicted segmentation mask and the ground truth. Let $P$ be the predicted mask and $G$ the ground truth mask:

\begin{equation}
\text{IoU}(P, G) = \frac{|P \cap G|}{|P \cup G|}
\end{equation}

\begin{equation}
\text{Dice}(P, G) = \frac{2 |P \cap G|}{|P| + |G|}
\end{equation}

\begin{equation}
\text{Accuracy}(P, G) = \frac{|P \cap G| + |\overline{P} \cap \overline{G}|}{|P \cup \overline{P}|}
\end{equation}

\begin{equation}
\text{Recall}(P, G) = \frac{|P \cap G|}{|G|}
\end{equation}

In \cite{deformable2017}, a universal dataset and a procedure were proposed in order to evaluate mosaic segmentation algorithms and compare them with past works. We explain them below for clarity.

Let \(N_{\text{gt}}\) denote the number of ground truth tiles, and \(N_{\text{pred}}\) denote the number of predicted regions. For each ground truth tile \(i\), let \(T_i\) represent the set of pixels belonging to that tile, and for each predicted region \(j\), let \(R_j\) represent the set of pixels in that region.

For each ground truth tile \(T_i\), let \(\text{region}_i\) be the predicted region \(R_j\) that maximizes the size of the intersection \(T_i \cap R_j\), i.e.,
\begin{equation}
    \text{region}_i = \arg\max_j |T_i \cap R_j|.
\end{equation}
Essentially, $\text{region}_i$ is the predicted tile (connected region) that overlaps the most with the ground truth tile $T_i$.

The Count Error is defined as
\begin{equation}
    \text{Count Error} = \frac{|N_{\text{gt}} - N_{\text{pred}}|}{N_{\text{gt}}}.
\end{equation}

The Precision metric is computed as the average, over all ground truth tiles, of the fraction of pixels in the best matching predicted region that also belong to the ground truth tile:
\begin{equation}
    \text{Precision} = \frac{1}{N_{\text{gt}}} \sum_{i=1}^{N_{\text{gt}}} \frac{|T_i \cap \text{region}_i|}{|\text{region}_i|},
\end{equation}
where \(|\text{region}_i|\) denotes the total number of pixels in the predicted region \(\text{region}_i\).

The Recall metric is computed as the average, over all ground truth tiles, of the fraction of pixels in the ground truth tile that are correctly predicted in the best matching predicted region:
\begin{equation}
    \text{Recall} = \frac{1}{N_{\text{gt}}} \sum_{i=1}^{N_{\text{gt}}} \frac{|T_i \cap \text{region}_i|}{|T_i|},
\end{equation}
where \(|T_i|\) is the total number of pixels in the ground truth tile \(T_i\).

Finally, the F-measure is the harmonic mean of Precision and Recall:
\begin{equation}
    \text{F-measure} = 2 \times \frac{\text{Precision} \times \text{Recall}}{\text{Precision} + \text{Recall}}.
\end{equation}

\section{Results}

Throughout this search for the optimal configuration of fine-tuning the SAM~2 model we found that the optimal hyperparameter values are as follows: a learning rate of \(10^{-5}\), a weight decay of \(10^{-5}\), a batch size of 4, and training for 9 epochs.

\begin{figure}[tb]
    \centerline{
        \begin{minipage}{\linewidth}
            \raggedright
            \includegraphics[width=\linewidth]{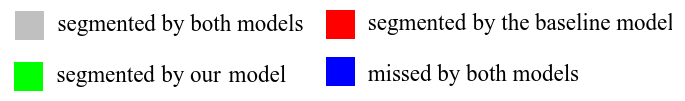}
        \end{minipage}
    }

    \vspace{0.3cm}

    \centerline{
        \begin{minipage}{0.9\linewidth}
            \centering
            \includegraphics[width=\linewidth]{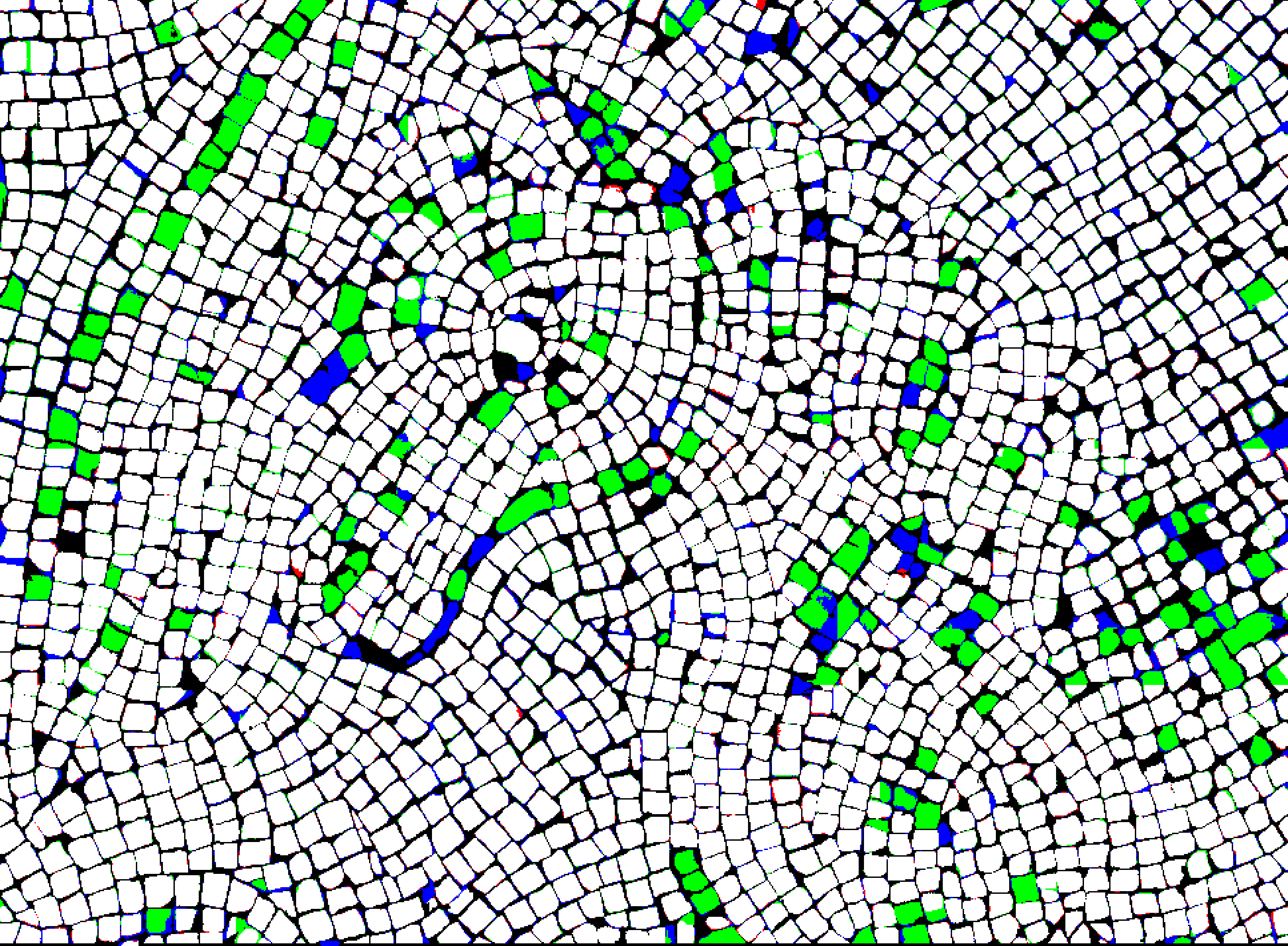}
        \end{minipage}
    }

    \caption{Example comparison of the baseline \texttt{SAM2\_base} and our fine-tuned \texttt{SAM2\_ft} model on the example image of Fig.~\ref{fig:mosaic_and_mask}. White pixels are correctly segmented by both models, green are correctly segmented only by \texttt{SAM2\_ft}, red only by \texttt{SAM2\_base}, while blue pixels are missed by both. Black pixels represent the background.}
    \label{fig:sample3-comparison}
\end{figure}

Table~\ref{tab:evaluation_results} summarizes the quantitative results of our testing, highlighting the improvements achieved by the fine-tuned models compared to the base checkpoints using the key metrics: IoU, Accuracy, Dice Coefficient and Recall. Both models were evaluated on our dataset, and the metrics were averaged across all images. The \texttt{SAM2\_ft} model outperforms \texttt{SAM2\_base} across all metrics. Fig.~\ref{fig:sample3-comparison} provides a visual comparison between the baseline SAM 2 and our fine-tuned model for the example image shown in Fig.~\ref{fig:mosaic_and_mask}, further demonstrating the improved performance of our model.

\begin{table}[tb]
\centering
\caption{Evaluation Results for the Baseline SAM~2 and our Fine-Tuned Model (in \%).}
\label{tab:evaluation_results}
\vspace{0.5em}
\renewcommand{\arraystretch}{1.4}
\begin{tabular}{|l||c|c|c|c|}
\hline
Model & IoU & Accuracy & Dice & Recall \\
\hline
\hline
\texttt{SAM2\_base} & 89 & 91 & 94 & 92 \\
\hline
\texttt{SAM2\_ft} & 91 & 93 & 95 & 96 \\
\hline
\end{tabular}
\end{table}

\subsubsection{Evaluation on Our Dataset}
Although the whole dataset proposed by \cite{deformable2017} was not accessible, the researchers kindly provided two available images (Museum with ID: 7 and University with ID: 11). We evaluate our model while adapting the findings of the previous works based on this limited subset. We acknowledge that a subset of only two images is insufficient to draw statistically significant conclusions regarding the model's segmentation performance and compare it to previous works. We address this by proposing our own state-of-the-art dataset of manually annotated mosaics, which we hope will help researchers in the future achieve even more accurate segmentations. Table~\ref{tab:our-dataset-results} presents segmentation results for different mosaic images using our model. It includes seven individual images as well as an average row. The values vary across the images, with precision ranging from the mid-70s to mid-90s, recall from the low 60s to low 80s, and F-measure from around 70 to nearly 87. The average values across all images are: $\text{Cnt} = 1.08$, $\text{Prec} = 86\%$, $\text{Rec} = 72\%$, and $\text{Fm} = 78\%$.

\begin{table}[tb]
\centering
\caption{Segmentation results on the dataset we propose (Prec, Rec, Fm in \%; Cnt is the count error).}
\small

\begin{tabular}{|c|c|c|c|c|}
\hline
\multirow{2}{*}{Image} & \multicolumn{4}{c|}{Our Method} \\
\cline{2-5}
& Cnt & Prec & Rec & Fm \\
\hline
Lizard          & 0.26 & 92 & 82 & 87 \\
Horse           & 4.89 & 84 & 60 & 70 \\
Tiger           & 1.08 & 76 & 67 & 71 \\
Duck            & 0.16 & 88 & 68 & 77 \\
Pattern 1       & 0.33 & 94 & 70 & 80 \\
Pattern 2       & 0.69 & 74 & 72 & 73 \\
Pattern 3       & 0.13 & 92 & 82 & 87 \\
\hline
Average         & 1.08 & 86 & 72 & 78 \\
\hline
\end{tabular}
\label{tab:our-dataset-results}
\end{table}

\begin{figure}[tb]
\centerline{\includegraphics[width=0.7\linewidth]{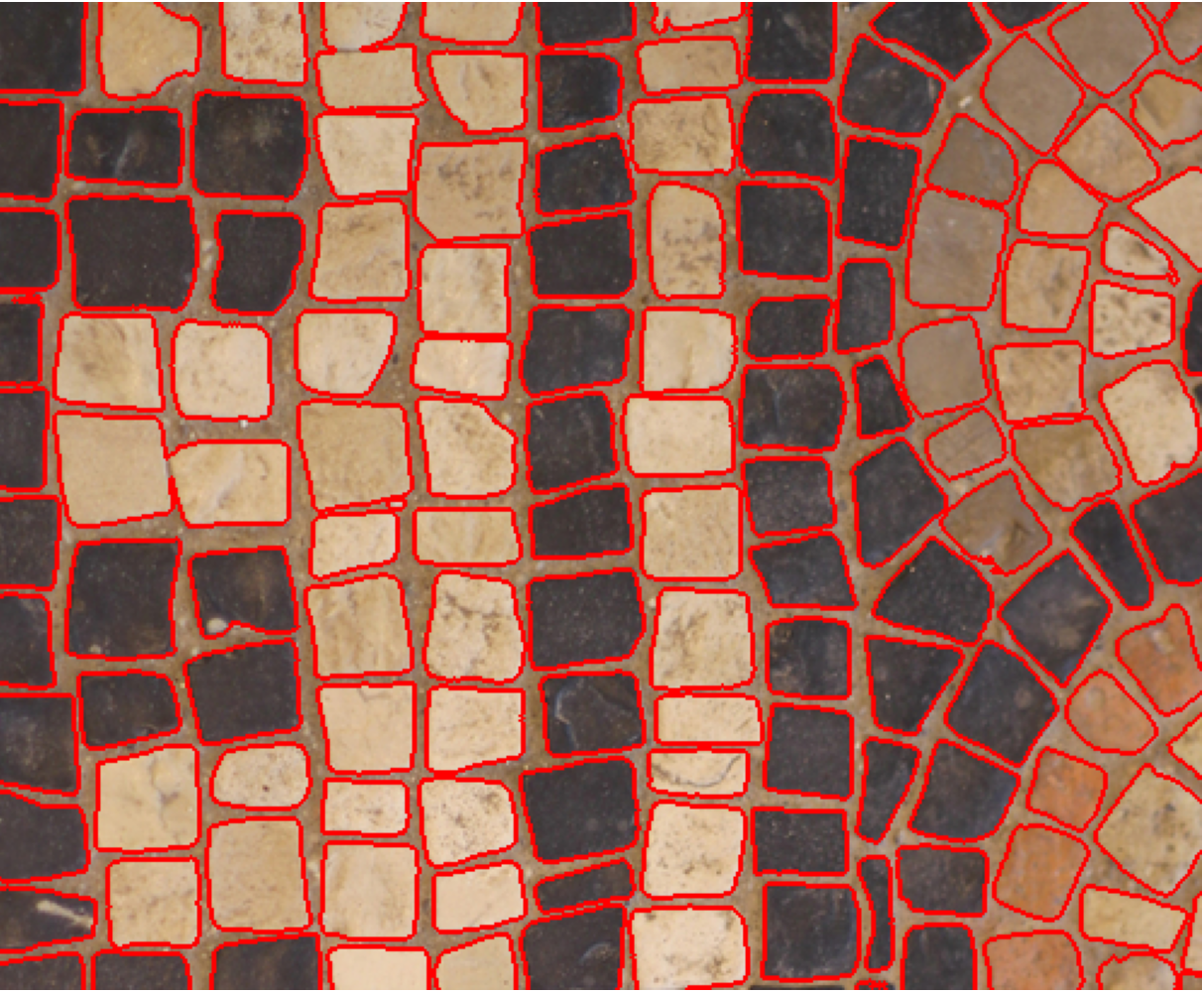}}
\caption{Segmentation of the Museum image with our method.}
\label{fig:museum}
\end{figure}

\subsubsection{Comparison to Previous Works}

In order to compare our model with previous works we evaluate it on the two images provided by the researchers \cite{deformable2017}. The results are organized in Table~\ref{tab:comparison-past-works} presenting segmentation performance across different images and methods, with metrics reported for each case. The fine-tuned SAM~2 method achieves the highest overall F-measure while also yielding the lowest Cnt values across the compared approaches.

\begin{table}[tb]
\caption{Segmentation Results on Image IDs 7 and 11 Across All Methods (Prec, Rec, Fm in \%; Cnt is the count error).}
\begin{center}
\renewcommand{\arraystretch}{1.2}

\begin{tabular}{|c|cccc|cccc|}
\hline
\multirow{2}{*}{ImageID} & \multicolumn{4}{c|}{Tessella-Oriented Segmentation} & \multicolumn{4}{c|}{Genetic Algorithms} \\
\cline{2-9}
& Cnt & Prec & Rec & Fm & Cnt & Prec & Rec & Fm \\
\hline
7  & 0.14 & 64 & 87 & 74 & 0.03 & 50 & 76 & 60 \\
11 & 0.90 & 63 & 78 & 70 & 0.03 & 46 & 67 & 55 \\
\hline
Avg & 0.52 & 64 & 83 & 72 & 0.03 & 48 & 72 & 58 \\
\hline
\end{tabular}

\vspace{1em}

\begin{tabular}{|c|cccc|cccc|}
\hline
\multirow{2}{*}{ImageID} & \multicolumn{4}{c|}{U-Net} & \multicolumn{4}{c|}{U-Net3} \\
\cline{2-9}
& Cnt & Prec & Rec & Fm & Cnt & Prec & Rec & Fm \\
\hline
7  & 0.30 & 65 & 80 & 73 & 0.25 & 75 & 90 & 82 \\
11 & 0.29 & 65 & 72 & 69 & 0.14 & 71 & 83 & 77 \\
\hline
Avg & 0.30 & 65 & 76 & 71 & 0.20 & 73 & 87 & 80 \\
\hline
\end{tabular}

\vspace{1em}

\begin{tabular}{|c|cccc|cccc|}
\hline
\multirow{2}{*}{ImageID} & \multicolumn{4}{c|}{SAM~2 (Pretrained)} & \multicolumn{4}{c|}{Fine-Tuned SAM~2} \\
\cline{2-9}
& Cnt & Prec & Rec & Fm & Cnt & Prec & Rec & Fm \\
\hline
7  & 12.38 & 62 & 75 & 68 & 0.01 & 76 & 81 & 78 \\
11 & 1.52  & 83 & 87 & 85 & 0.03 & 89 & 86 & 87 \\
\hline
Avg & 6.95 & 72 & 81 & 76 & 0.02 & 82 & 83 & 83 \\
\hline
\end{tabular}

\label{tab:comparison-past-works}
\end{center}
\end{table}

\section{Discussion}

This work experimented with the SAM~2 model and how it performed in the task of automatic mosaic segmentation to separate the individual tiles. We used transfer learning \cite{bell2025} and achieved highly accurate segmentation on a newly annotated dataset of mosaics. Both visual comparisons and evaluation metrics illustrated a significant improvement in comparison to the baseline model of SAM~2 and previous works. We provide a benchmark dataset\footnote{\url{https://huggingface.co/datasets/ckapelonis02/mosaics-images}}, the accompanying code\footnote{\url{https://github.com/ckapelonis02/mosaic-sam2}}, and a fine-tuned SAM~2 model\footnote{\url{https://huggingface.co/ckapelonis02/SAM2-mosaic}} to facilitate future comparisons and experiments on mosaic tesserae segmentation.

Table~\ref{tab:evaluation_results} and Fig.~\ref{fig:sample3-comparison} together illustrate the improvement achieved by fine-tuning SAM~2. Table~\ref{tab:evaluation_results} shows that the fine-tuned model (\texttt{SAM2\_ft}) outperforms the baseline (\texttt{SAM2\_base}) across all metrics, with higher IoU, Accuracy, Dice, and Recall values. Figure~\ref{fig:sample3-comparison} visually confirms this: \texttt{SAM2\_ft} captures more correct pixels (green) that the baseline misses (red), while maintaining most of the correctly segmented regions shared by both models (white). Overall, the fine-tuned model produces more accurate and complete segmentations.

Table~\ref{tab:comparison-past-works} demonstrates the performance comparison between previous methods and our work. The Tessella-Oriented Segmentation (TOS) \cite{WT2008} method had a very strong Fm value of 72\%, albeit showing a weak Cnt value of 0.52. The method based on Genetic Algorithms (GA) \cite{deformable2017} lacked in the Fm metric, displaying a poor performance of 58\%, yet achieved a Cnt of 0.03. The low percentage of Fm suggested that these predicted regions might be misaligned or inaccurate. The fine-tuned U-Net \cite{unet2020} model's results showed a Cnt of 0.30 and Fm of 71\%, proving that DL is the future of efficiently solving this task of mosaics segmentation. Further proof of the above was given by the Mo.Se. U-Net3 \cite{mose2021} model, with a Cnt of 0.20 and Fm of 80\%. Our fine-tuned SAM~2 model achieved a Cnt of 0.02 and Fm of 83\%, indicating that our approach performed reliably on the given testing subset.

One of the biggest limitations we faced was the 30-hours quota by Kaggle\footnote{\url{https://www.kaggle.com/}}. At the same time, the GPU and RAM provided by the platform were not as powerful as advised by the Meta AI team in the training instructions\footnote{\url{https://github.com/facebookresearch/sam2}} by a great margin (e.g., GPU NVIDIA Tesla P100 vs A100). The manual annotation process was restricted to a small set of images, thereby not allowing the model to reach the full potential due to data scarcity through the training process. Expanding the mosaic dataset, especially with images provided by museums, could substantially improve the robustness and generalization abilities of the model.

While the task of automatically segmenting mosaics seems conceptually easy, we encountered many unexpected problems throughout. Similarly to any ML problem, more computational power and time abundance would accelerate the process and produce better results. Additionally, a future update can include exploring alternative segmentation models with stronger transfer learning capabilities (e.g., \cite{qin2022}), adapting the SAM~2 architecture to fit the given task, while being more lightweight to enable execution on less powerful hardware, and refining post-processing techniques, such as adjusting inaccurate segmentations by merging fragmented tesserae or better separating overlapping ones. Future work can also explore extensions to 3D mosaics to complement digital restoration tools. Finally, we could provide further benchmarks for comparison, encorporating statistical examination for more precise analysis.

\section{Conclusion}

We developed an accurate, fully automatic segmentation model for mosaic images. Quantitative results on a newly annotated dataset showed substantial improvement over the baseline SAM~2 model and previous works, while qualitative results, such as visual comparisons on real mosaics, further confirm that our fine-tuned model outperforms previous approaches and offers a promising step toward automating tesserae segmentation.

\section*{Acknowledgment}

This study was supported by the framework of the Action ``Interreg VI-A \guillemotleft Greece-Cyprus 2021-2027\guillemotright'', which is implemented through the special service ``Interreg 2021 - 2027'' of the Greek Ministry of National Economy and Finance and co-financed by the European Union (ERDF) and by National Resources of Greece and Cyprus (Project ID (MIS: 6006535)).

\end{document}